\documentclass{article} 
\usepackage{bm}
\usepackage{nips14submit_e,times}
\usepackage{url}
\usepackage{amssymb}
\usepackage{amsthm}
\usepackage[numbers]{natbib}
\usepackage{enumitem}
\usepackage{amsmath}  
\usepackage{array}

\usepackage{algorithm}
\usepackage{algorithmic}

\usepackage{graphicx}

\title{Conditional Generative Adversarial Nets}

\author{
Mehdi Mirza \\
D\'epartement d'informatique et de recherche op\'erationnelle\\
Universit\'e de Montr\'eal\\
Montr\'eal, QC H3C 3J7 \\
\texttt{mirzamom@iro.umontreal.ca}
\AND
Simon Osindero \\
Flickr / Yahoo Inc. \\
San Francisco, CA 94103 \\
\texttt{osindero@yahoo-inc.com} \\
}

\nipsfinalcopy 

\begin{document}

\maketitle

\begin{abstract}
\textbf{Generative Adversarial Nets} \cite{Goodfellow-et-al-NIPS2014-small} were recently introduced as a novel
way to train generative models. In this work we introduce the conditional
version of generative adversarial nets, which can be constructed by simply feeding
the data, ${y}$, we wish to condition on to both the generator and discriminator.
We show that this model can generate MNIST digits conditioned on class labels.
We also illustrate how this model could be used to learn a multi-modal model, and provide
preliminary examples of an application to image tagging in which we demonstrate
how this approach can generate descriptive tags which are not part of training labels.
\end{abstract}

\section{Introduction}

Generative adversarial nets were recently introduced as an alternative framework for training
generative models in order to sidestep the difficulty of approximating many intractable probabilistic
computations.

Adversarial nets have the advantages that Markov chains
are never needed, only backpropagation is used to obtain gradients,
no inference is required during learning,
and a wide variety of factors and interactions can easily be incorporated into the model.

Furthermore, as demonstrated in \cite{Goodfellow-et-al-NIPS2014-small},
it can produce state of the art log-likelihood estimates and realistic samples.

In an unconditioned generative model, there is no control on modes of the data being generated.
However, by conditioning the model on additional information it is possible to direct the data
generation process. Such conditioning could be based on class labels, on some part of data
for inpainting like \cite{goodfellow2013multi}, or even on data from different modality.

In this work we show how can we construct the conditional adversarial net. And for empirical
results we demonstrate two set of experiment. One on MNIST digit data set conditioned on class
labels and one on MIR Flickr 25,000 dataset \cite{huiskes08} for multi-modal learning.

\section{Related Work}

\subsection{Multi-modal Learning For Image Labelling}

Despite the many recent successes of supervised neural networks  (and convolutional networks in particular) \cite{Krizhevsky-2012, szegedy2014going},
it remains challenging to scale such models to accommodate an extremely large number of predicted output categories. A second issue is
that much of the work to date has focused on learning one-to-one mappings from input to output. However, many
interesting problems are more naturally thought of as a probabilistic one-to-many mapping. For instance in the case of image labeling
there may be many different tags that could appropriately applied to a given image, and different (human) annotators may
use different (but typically synonymous or related) terms to describe the same image.

One way to help address the first issue is to leverage additional information from other modalities:
for instance, by using natural language corpora to learn a vector representation for labels in which geometric relations are semantically meaningful.
When making predictions in such spaces, we benefit from the fact that when prediction errors
we are still often `close' to the truth (e.g. predicting 'table' instead of 'chair'), and also from the fact that we can naturally make predictive generalizations
to labels that were not seen during training time. Works such as \cite{frome2013devise} have shown that even a
simple linear mapping from image feature-space to word-representation-space can yield improved classification performance.

One way to address the second problem is to use a conditional probabilistic generative model, the input is taken to be
the conditioning variable and the one-to-many mapping is instantiated as a conditional predictive distribution.


\cite{Srivastava+Salakhutdinov-NIPS2012-small} take a similar approach to this problem, and train a multi-modal
Deep Boltzmann Machine on the MIR Flickr 25,000 dataset as we do in this work.

Additionally, in \cite{kiros2013multimodal} the authors show how to train a supervised multi-modal neural language model, and they are able to
generate descriptive sentence for images.

\section{Conditional Adversarial Nets}
\subsection{Generative Adversarial Nets}
Generative adversarial nets were recently introduced as a novel way to train a generative model. They consists
of two `adversarial' models: a generative model ${G}$ that captures the data distribution, and a discriminative model ${D}$
that estimates the probability that a sample came from the training data rather than ${G}$.
Both ${G}$ and ${D}$ could be a non-linear mapping function, such as a multi-layer perceptron.

To learn a generator distribution ${p_g}$ over data data ${\bm{x}}$, the generator builds a mapping function
from a prior noise distribution ${p_z(z)}$ to data space as ${G(z;\theta_g)}$. And the discriminator, ${D(x; \theta_d)}$, outputs
a single scalar representing the probability that ${\bm{x}}$ came form training data rather than ${p_g}$.

${G}$ and ${D}$ are both trained simultaneously: we adjust parameters for ${G}$ to minimize ${log(1-D(G(\bm{z}))}$ and
adjust parameters for $D$ to minimize ${logD(X)}$, as if they are following the two-player min-max game with value function ${V(G, D)}$:

\begin{equation}
\label{eq:minimaxgame-definition}
\min_G \max_D V(D, G) = \mathbb{E}_{\bm{x} \sim p_{\text{data}}(\bm{x})}[\log D(\bm{x})] + \mathbb{E}_{\bm{z} \sim p_z(\bm{z})}[\log (1 - D(G(\bm{z})))].
\end{equation}

\subsection{Conditional Adversarial Nets}

Generative adversarial nets can be extended to a conditional model if
both the generator and discriminator are conditioned on some extra information ${\bm{y}}$.
${\bm{y}}$ could be any kind of auxiliary information, such as class labels or
data from other modalities. We can perform the conditioning by feeding ${\bm{y}}$ into the both the discriminator
and generator as additional input layer.

In the generator the prior input noise $p_{\bm{z}}(\bm{z})$,
and ${\bm{y}}$ are combined in joint hidden representation, and the adversarial training framework allows for
considerable flexibility in how this hidden representation is composed. \footnote{For now we simply have the conditioning
input and prior noise as inputs to a single hidden layer of a MLP, but one could imagine using higher order interactions
allowing for complex generation mechanisms that would be extremely difficult to work with in a traditional generative framework.}

In the discriminator ${\bm{x}}$ and
${\bm{y}}$ are presented as inputs and to a discriminative function (embodied again by a MLP in this case).

The objective function of a two-player minimax game would be as Eq \ref{eq:minimaxgame-definition-conditioned}
\begin{equation}
\label{eq:minimaxgame-definition-conditioned}
\min_G \max_D V(D, G) = \mathbb{E}_{\bm{x} \sim p_{\text{data}}(\bm{x})}[\log D(\bm{x} | \bm{y})] + \mathbb{E}_{\bm{z} \sim p_z(\bm{z})}[\log (1 - D(G(\bm{z} | \bm{y})))].
\end{equation}

\begin{figure}[h]
\centering
    \includegraphics[width=0.9\textwidth]{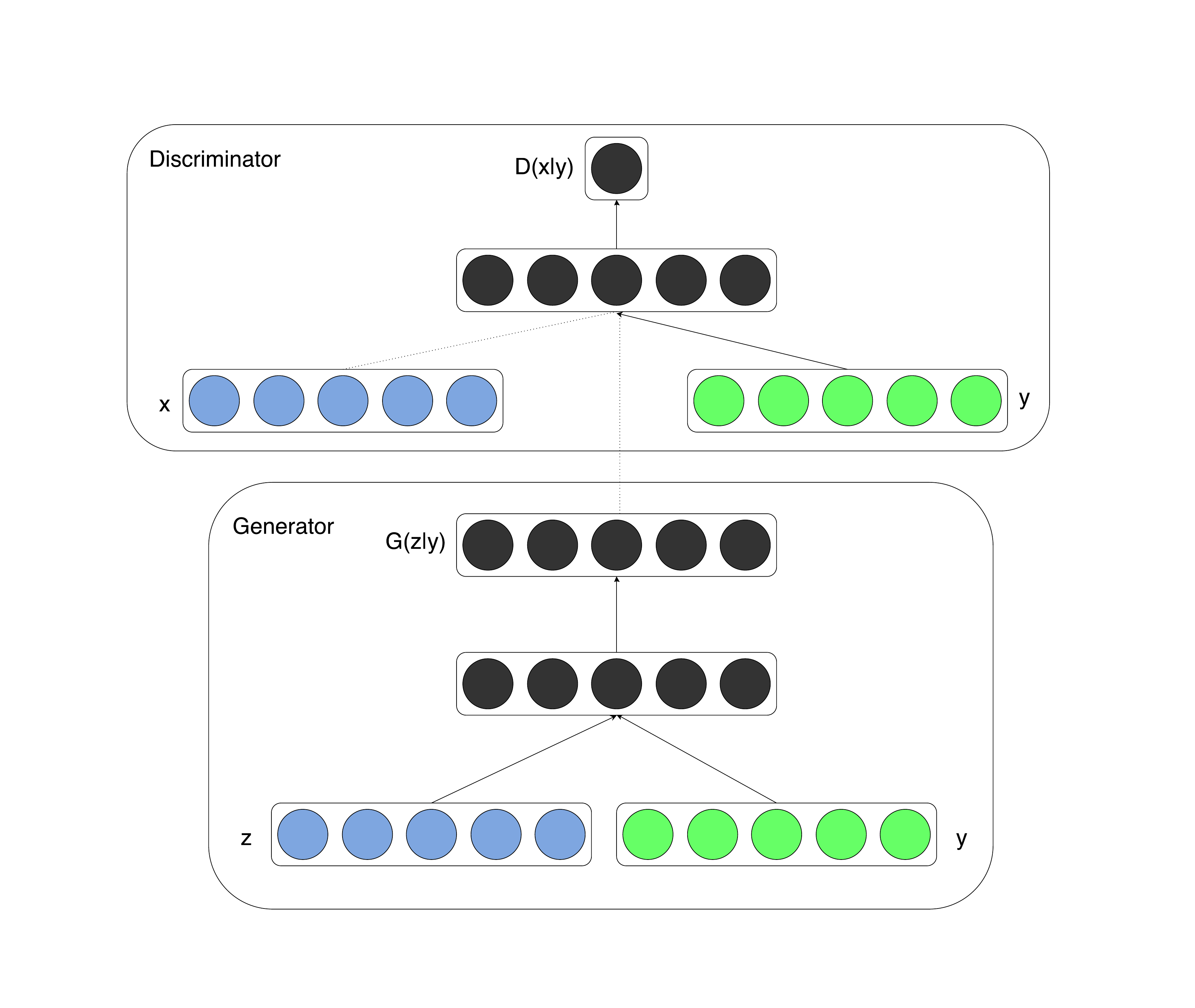}
    \caption{\small Conditional adversarial net}
\label{fig:diagram}
\end{figure}

Fig \ref{fig:diagram} illustrates the structure of a simple conditional adversarial net.

\section{Experimental Results}
\subsection{Unimodal}
We trained a conditional adversarial net on MNIST images conditioned
on their class labels, encoded as one-hot vectors.

In the generator net, a noise prior ${\bm{z}}$ with dimensionality 100 was drawn from a uniform distribution within the unit hypercube.
Both  ${\bm{z}}$ and ${\bm{y}}$ are mapped to hidden layers with Rectified Linear
Unit (ReLu) activation \cite{glorot2011deep, Jarrett-ICCV2009-small}, with
layer sizes 200 and 1000 respectively, before both being mapped to second, combined hidden ReLu layer of dimensionality 1200.
We then have a final sigmoid unit layer as our output for generating the 784-dimensional MNIST samples.

The discriminator maps ${\bm{x}}$ to a maxout \cite{Goodfellow_maxout_2013} layer with 240
units and 5 pieces, and  ${\bm{y}}$ to a maxout layer with 50 units and 5 pieces. Both of the hidden
layers mapped to a joint maxout layer with 240 units and 4 pieces before being fed to
the sigmoid layer. (The precise architecture of the discriminator is not critical as long as it has
sufficient power; we have found that maxout units are typically well suited to the task.)

The model was trained using stochastic gradient decent with mini-batches of size 100
and initial learning rate of $0.1$
which was exponentially decreased down to $.000001$ with decay factor of $1.00004$.
Also momentum was used with initial value of $.5$ which was increased up to $0.7$.
Dropout \cite{Hinton-et-al-arxiv2012} with probability of 0.5 was applied to both the generator and discriminator.
And best estimate of log-likelihood on the validation set was used as stopping point.

Table \ref{table:parzen} shows Gaussian Parzen window log-likelihood estimate for the MNIST dataset test data.
1000 samples were drawn from each 10 class and a Gaussian Parzen window  was fitted to these samples.
We then estimate the log-likelihood of the test set using the Parzen window distribution.
(See \cite{Goodfellow-et-al-NIPS2014-small} for more details of how this estimate is constructed.)

The conditional adversarial net results that we present are comparable with some other network based, but
are outperformed by several other approaches -- including non-conditional adversarial nets.
We present these results more as a proof-of-concept than as demonstration of efficacy, and believe that
with further exploration of hyper-parameter space and architecture that the conditional model should match
or exceed the non-conditional results.

\begin{figure}[h]
	\centering
	    \includegraphics[width=0.9\textwidth]{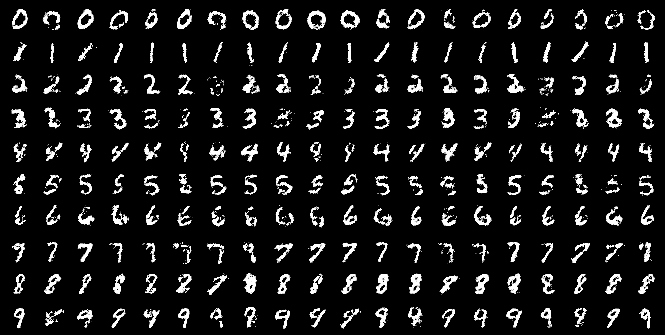}
	    \caption{\small Generated MNIST digits, each row conditioned on one label}
	\label{fig:mnist}
\end{figure}

Fig \ref{fig:mnist} shows some of the generated samples. Each row is
conditioned on one label and each column is a different generated sample.

\begin{table}
	\centering
	\begin{tabular}{c|c}
	Model & MNIST  \\
	\hline
	DBN~\citep{Bengio-et-al-ICML2013} & $138 \pm 2$  \\
	Stacked CAE~\citep{Bengio-et-al-ICML2013} & $121 \pm 1.6$ \\
	Deep GSN~\citep{Bengio-et-al-ICML-2014} & $214 \pm 1.1$ \\
	Adversarial nets & $225 \pm 2$ \\
	Conditional adversarial nets & $132 \pm 1.8$
	\end{tabular}
	\caption{\small
	Parzen window-based log-likelihood estimates for MNIST. We followed the same procedure as
	\cite{Goodfellow-et-al-NIPS2014-small} for computing these values.}
	\label{table:parzen}
\end{table}

\subsection{Multimodal}
Photo sites such as Flickr are a rich source of labeled data in the form of images and
their associated user-generated metadata (UGM) --- in particular user-tags.

User-generated metadata differ from more `canonical' image labelling schems in that they are typically
more descriptive, and are semantically much closer to how humans describe images with natural language
rather than just identifying the objects present in an image. Another aspect of UGM is that synoymy is
prevalent and different users may use different vocabulary to describe the same concepts --- consequently,
having an efficient way to normalize these labels becomes important. Conceptual word embeddings \cite{mikolov-et-al-iclr2013}
can be very useful here since related concepts end up being represented by similar vectors.

In this section we demonstrate automated tagging of images, with multi-label predictions,
using conditional adversarial nets to generate a (possibly multi-modal) distribution
of tag-vectors conditional on image features.


For image features we pre-train a convolutional model similar to the one from \cite{Krizhevsky-2012}
on the full ImageNet dataset with 21,000 labels \cite{RussakovskyFeiFei}.
We use the output of the last fully connected layer with 4096 units as image representations.

For the world representation we first gather a corpus of text from concatenation of user-tags, titles and descriptions from
YFCC100M \footnote{Yahoo Flickr Creative Common 100M \url{http://webscope.sandbox.yahoo.com/catalog.php?datatype=i&did=67}.}
dataset metadata. After pre-processing and cleaning of the text we trained a
skip-gram model \cite{mikolov-et-al-iclr2013} with word vector size of 200. And we omitted any word appearing
less than 200 times from the vocabulary,  thereby ending up with a dictionary of size 247465.

We keep the convolutional model and the language model fixed during training of the adversarial net.
And leave the experiments when we even backpropagate through these models as future work.

For our experiments we use MIR Flickr 25,000 dataset \cite{huiskes08}, and extract the image and
tags features using the convolutional model and language model we described above.
Images without any tag were omitted from our experiments and annotations were treated as extra tags.
The first 150,000 examples were used as training set.
Images with multiple tags were repeated inside the training set once for each associated tag.

For evaluation, we generate 100 samples for each image and find top 20 closest words
using cosine similarity of vector representation of the words in the vocabulary to each sample.
Then we select the top 10 most common words among all 100 samples.
Table \ref{table:samples} shows some samples of the user
assigned tags and annotations along with the generated tags.


The best working model's generator receives Gaussian noise of size 100 as noise prior and maps it to 500 dimension ReLu layer.
And maps 4096 dimension image feature vector to 2000 dimension ReLu hidden layer.
Both of these layers are mapped to a joint representation of 200 dimension linear layer which would output the generated word vectors.

The discriminator is consisted of 500 and 1200 dimension ReLu hidden layers for word vectors
and image features respectively and maxout layer with 1000 units and 3 pieces as the join layer which is finally fed to
the one single sigmoid unit.

The model was trained using stochastic gradient decent with mini-batches of size 100
and initial learning rate of $0.1$
which was exponentially decreased down to $.000001$ with decay factor of $1.00004$.
Also momentum was used with initial value of $.5$ which was increased up to $0.7$.
Dropout with probability of 0.5 was applied to both the generator and discriminator.

The hyper-parameters and architectural choices were obtained by cross-validation
 and a mix of random grid search and manual selection (albeit over a somewhat limited search space.)

\begin{table}[h]
\begin{tabular} {l | b{3.7cm} | b{3.7cm}}
  & User tags + annotations & Generated tags \\
 \hline
 \includegraphics[width=0.33\textwidth, height=0.12\textheight]{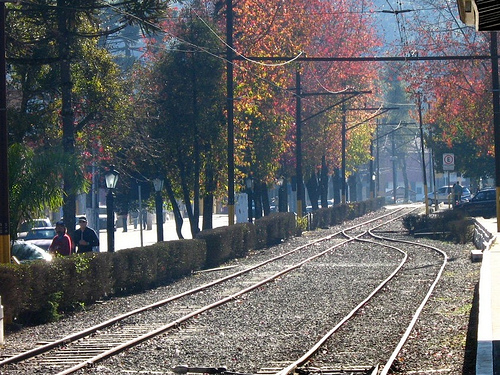} & montanha, trem, inverno, frio, people, male, plant life, tree, structures, transport, car & taxi, passenger, line, transportation, railway station, passengers, railways, signals, rail, rails \\
 \includegraphics[width=0.33\textwidth, height=0.12\textheight]{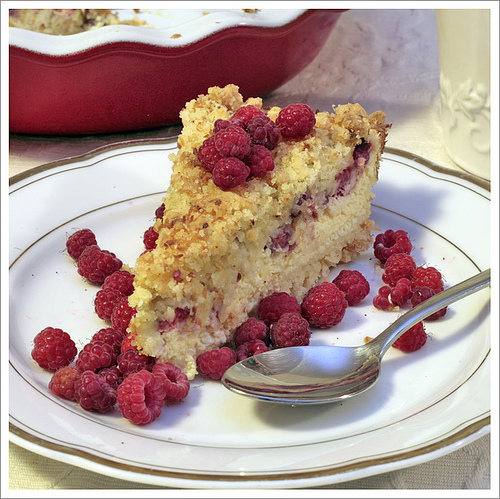} & food, raspberry, delicious, homemade & chicken, fattening, cooked, peanut, cream, cookie, house made, bread, biscuit, bakes \\

 \includegraphics[width=0.33\textwidth, height=0.12\textheight]{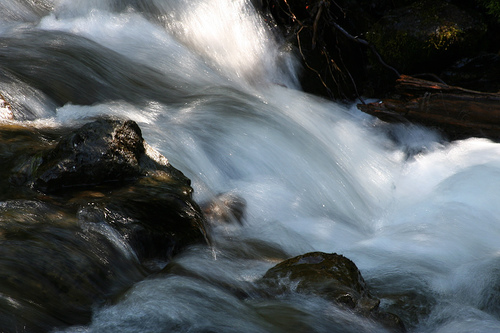} & water, river & creek, lake, along, near, river, rocky, treeline, valley, woods, waters \\

 \includegraphics[width=0.33\textwidth, height=0.12\textheight]{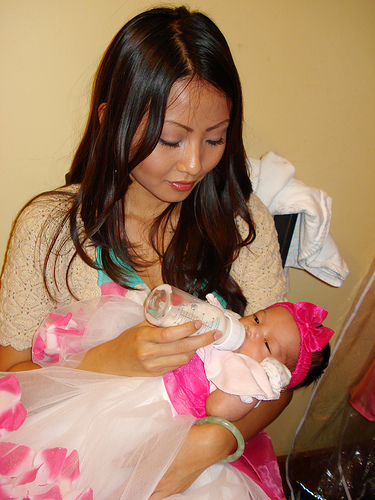} & people, portrait, female, baby, indoor & love, people, posing, girl, young, strangers, pretty, women, happy, life \\
\end{tabular}
\label{table:samples}
\caption{Samples of generated tags}
\end{table}

\section{Future Work}
The results shown in this paper are extremely preliminary, but they demonstrate the potential of
conditional adversarial nets and show promise for interesting and useful applications.

In future explorations between now and the workshop we expect to present more sophisticated models,
as well as a more detailed and thorough analysis of their performance and characteristics.


Also, in the current experiments we only use each tag individually.
But by using multiple tags at the same time (effectively posing generative problem as one of `set generation')
we hope to achieve better results.

Another obvious direction left for future work is to construct a joint training scheme to learn the language model. Works such as
\cite{kiros2013multimodal} has shown that we can learn a language model for suited for the specific task.

\subsubsection*{Acknowledgments}
This project was developed in
Pylearn2~\citep{goodfellow2013pylearn2} framework, and we would like to thank Pylearn2 developers. 
We also like to thank Ian Goodfellow for helpful discussion during his affiliation at University of Montreal.
The authors gratefully acknowledge the support from the Vision \& Machine Learning, and Production
Engineering teams at Flickr (in alphabetical order: Andrew Stadlen, Arel Cordero, Clayton Mellina,
Cyprien Noel, Frank Liu, Gerry Pesavento, Huy Nguyen, Jack Culpepper, John Ko, Pierre Garrigues,
Rob Hess, Stacey Svetlichnaya, Tobi Baumgartner, and Ye Lu).

\small{
\bibliography{strings,strings-shorter,ml}

\begin{thebibliography}{}

\bibitem[Bengio {\em et~al.}(2013)Bengio, Mesnil, Dauphin, and
  Rifai]{Bengio-et-al-ICML2013}
Bengio, Y., Mesnil, G., Dauphin, Y., and Rifai, S. (2013).
\newblock Better mixing via deep representations.
\newblock In {\em ICML'2013\/}.

\bibitem[Bengio {\em et~al.}(2014)Bengio, Thibodeau-Laufer, Alain, and
  Yosinski]{Bengio-et-al-ICML-2014}
Bengio, Y., Thibodeau-Laufer, E., Alain, G., and Yosinski, J. (2014).
\newblock Deep generative stochastic networks trainable by backprop.
\newblock In {\em Proceedings of the 30th International Conference on Machine
  Learning (ICML'14)\/}.

\bibitem[Frome {\em et~al.}(2013)Frome, Corrado, Shlens, Bengio, Dean, Mikolov,
  {\em et~al.}]{frome2013devise}
Frome, A., Corrado, G.~S., Shlens, J., Bengio, S., Dean, J., Mikolov, T., {\em
  et~al.} (2013).
\newblock Devise: A deep visual-semantic embedding model.
\newblock In {\em Advances in Neural Information Processing Systems\/}, pages
  2121--2129.

\bibitem[Glorot {\em et~al.}(2011)Glorot, Bordes, and Bengio]{glorot2011deep}
Glorot, X., Bordes, A., and Bengio, Y. (2011).
\newblock Deep sparse rectifier neural networks.
\newblock In {\em International Conference on Artificial Intelligence and
  Statistics\/}, pages 315--323.

\bibitem[Goodfellow {\em et~al.}(2013a)Goodfellow, Mirza, Courville, and
  Bengio]{goodfellow2013multi}
Goodfellow, I., Mirza, M., Courville, A., and Bengio, Y. (2013a).
\newblock Multi-prediction deep boltzmann machines.
\newblock In {\em Advances in Neural Information Processing Systems\/}, pages
  548--556.

\bibitem[Goodfellow {\em et~al.}(2013b)Goodfellow, Warde-Farley, Mirza,
  Courville, and Bengio]{Goodfellow_maxout_2013}
Goodfellow, I.~J., Warde-Farley, D., Mirza, M., Courville, A., and Bengio, Y.
  (2013b).
\newblock Maxout networks.
\newblock In {\em ICML'2013\/}.

\bibitem[Goodfellow {\em et~al.}(2013c)Goodfellow, Warde-Farley, Lamblin,
  Dumoulin, Mirza, Pascanu, Bergstra, Bastien, and
  Bengio]{goodfellow2013pylearn2}
Goodfellow, I.~J., Warde-Farley, D., Lamblin, P., Dumoulin, V., Mirza, M.,
  Pascanu, R., Bergstra, J., Bastien, F., and Bengio, Y. (2013c).
\newblock Pylearn2: a machine learning research library.
\newblock {\em arXiv preprint arXiv:1308.4214\/}.

\bibitem[Goodfellow {\em et~al.}(2014)Goodfellow, Pouget-Abadie, Mirza, Xu,
  Warde-Farley, Ozair, Courville, and Bengio]{Goodfellow-et-al-NIPS2014-small}
Goodfellow, I.~J., Pouget-Abadie, J., Mirza, M., Xu, B., Warde-Farley, D.,
  Ozair, S., Courville, A., and Bengio, Y. (2014).
\newblock Generative adversarial nets.
\newblock In {\em NIPS'2014\/}.

\bibitem[Hinton {\em et~al.}(2012)Hinton, Srivastava, Krizhevsky, Sutskever,
  and Salakhutdinov]{Hinton-et-al-arxiv2012}
Hinton, G.~E., Srivastava, N., Krizhevsky, A., Sutskever, I., and
  Salakhutdinov, R. (2012).
\newblock Improving neural networks by preventing co-adaptation of feature
  detectors.
\newblock Technical report, arXiv:1207.0580.

\bibitem[Huiskes and Lew(2008)Huiskes and Lew]{huiskes08}
Huiskes, M.~J. and Lew, M.~S. (2008).
\newblock The mir flickr retrieval evaluation.
\newblock In {\em MIR '08: Proceedings of the 2008 ACM International Conference
  on Multimedia Information Retrieval\/}, New York, NY, USA. ACM.

\bibitem[Jarrett {\em et~al.}(2009)Jarrett, Kavukcuoglu, Ranzato, and
  {LeCun}]{Jarrett-ICCV2009-small}
Jarrett, K., Kavukcuoglu, K., Ranzato, M., and {LeCun}, Y. (2009).
\newblock What is the best multi-stage architecture for object recognition?
\newblock In {\em ICCV'09\/}.

\bibitem[Kiros {\em et~al.}(2013)Kiros, Zemel, and
  Salakhutdinov]{kiros2013multimodal}
Kiros, R., Zemel, R., and Salakhutdinov, R. (2013).
\newblock Multimodal neural language models.
\newblock In {\em Proc. NIPS Deep Learning Workshop\/}.

\bibitem[Krizhevsky {\em et~al.}(2012)Krizhevsky, Sutskever, and
  Hinton]{Krizhevsky-2012}
Krizhevsky, A., Sutskever, I., and Hinton, G. (2012).
\newblock {ImageNet} classification with deep convolutional neural networks.
\newblock In {\em Advances in Neural Information Processing Systems 25
  (NIPS'2012)\/}.

\bibitem[Mikolov {\em et~al.}(2013)Mikolov, Chen, Corrado, and
  Dean]{mikolov-et-al-iclr2013}
Mikolov, T., Chen, K., Corrado, G., and Dean, J. (2013).
\newblock Efficient estimation of word representations in vector space.
\newblock In {\em International Conference on Learning Representations:
  Workshops Track\/}.

\bibitem[Russakovsky and Fei-Fei(2010)Russakovsky and
  Fei-Fei]{RussakovskyFeiFei}
Russakovsky, O. and Fei-Fei, L. (2010).
\newblock Attribute learning in large-scale datasets.
\newblock In {\em European Conference of Computer Vision (ECCV), International
  Workshop on Parts and Attributes\/}, Crete, Greece.

\bibitem[Srivastava and Salakhutdinov(2012)Srivastava and
  Salakhutdinov]{Srivastava+Salakhutdinov-NIPS2012-small}
Srivastava, N. and Salakhutdinov, R. (2012).
\newblock Multimodal learning with deep boltzmann machines.
\newblock In {\em NIPS'2012\/}.

\bibitem[Szegedy {\em et~al.}(2014)Szegedy, Liu, Jia, Sermanet, Reed, Anguelov,
  Erhan, Vanhoucke, and Rabinovich]{szegedy2014going}
Szegedy, C., Liu, W., Jia, Y., Sermanet, P., Reed, S., Anguelov, D., Erhan, D.,
  Vanhoucke, V., and Rabinovich, A. (2014).
\newblock Going deeper with convolutions.
\newblock {\em arXiv preprint arXiv:1409.4842\/}.

\end{thebibliography}
\bibliographystyle{natbib}}

\end{document}